\def\year{2021}\relax
\documentclass[letterpaper]{article} % DO NOT CHANGE THIS
\usepackage{aaai21}  % DO NOT CHANGE THIS
\usepackage{times}  % DO NOT CHANGE THIS
\usepackage{helvet} % DO NOT CHANGE THIS
\usepackage{courier}  % DO NOT CHANGE THIS
\usepackage[hyphens]{url}  % DO NOT CHANGE THIS
\usepackage{graphicx} % DO NOT CHANGE THIS
\usepackage{natbib}  % DO NOT CHANGE THIS AND DO NOT ADD ANY OPTIONS TO IT
\usepackage{caption} % DO NOT CHANGE THIS AND DO NOT ADD ANY OPTIONS TO IT
\usepackage{amssymb}
\usepackage{algorithm}
\usepackage{algorithmic}
\usepackage{amsmath}
\usepackage{subfigure}

\usepackage{multirow}
\usepackage{graphicx}
\usepackage{epstopdf}

\title{Cascade Network with Guided Loss and Hybrid Attention for Finding Good Correspondences}
\author{
    Zhi Chen, Fan Yang, Wenbing Tao\textsuperscript{}\thanks{Corresponding author.}\\
}
\affiliations{
    National Key Laboratory of Science and Technology on Multispectral Information Processing \\
    School of Artifical Intelligence and Automation, Huazhong University of Science and Technology, China\\
    \{hust\_zhichen, hust\_fanyang, wenbingtao\}@hust.edu.cn 

}
\begin{document}

\maketitle

\begin{abstract}
Finding good correspondences is a critical prerequisite in many feature based tasks. Given a putative correspondence set of an image pair, we propose a neural network which finds correct correspondences by a binary-class classifier and estimates relative pose through classified correspondences. First, we analyze that due to the imbalance in the number of correct and wrong correspondences, the loss function has a great impact on the classification results. Thus, we propose a new Guided Loss that can directly use evaluation criterion (Fn-measure) as guidance to dynamically adjust the objective function during training. We theoretically prove that the perfect negative correlation between the Guided Loss and Fn-measure, so that the network is always trained towards the direction of increasing Fn-measure to maximize it. We then propose a hybrid attention block to extract feature, which integrates the Bayesian attentive context normalization (BACN) and channel-wise attention (CA). BACN can mine the prior information to better exploit global context and CA can capture complex channel context to enhance the channel awareness of the network. Finally, based on our Guided Loss and hybrid attention block, a cascade network is designed to gradually optimize the result for more superior performance. Experiments have shown that our network achieves the state-of-the-art performance on benchmark datasets. Our code will be available in https://github.com/wenbingtao/GLHA.
\end{abstract}

\label{introduction_section}
\begin{figure}[t]
	\centering
	\includegraphics[width=1\columnwidth]{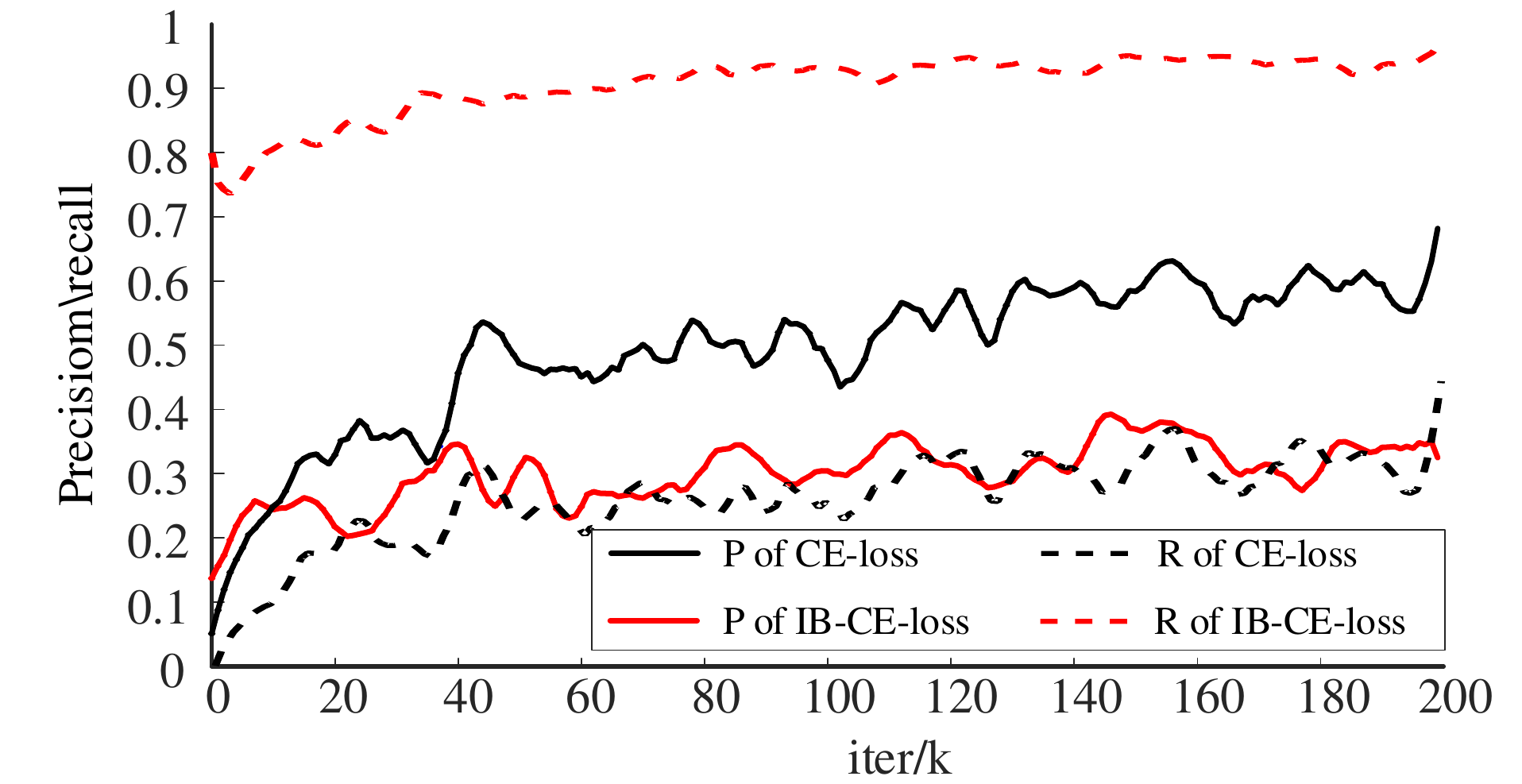}
	\caption{The training curves of different loss functions with same network on YFCC100M$\&$SUN3D dataset. In this dataset, the number ratio of positive and negative samples is about $1:10$. The imbalance of precision and recall occurs on the network using the above two loss functions.}\label{introduction}
	\label{fig1}
\end{figure}

\section{Introduction}
\noindent Two view geometry estimation, i.e., establishing reliable correspondences and estimating relative pose between an image pair, is the fundamental component of many tasks in computer vision, such as Structure from Motion (SfM) \cite{schonberger2016structure,snavely2006photo}, simultaneous localization and mapping (SLAM) \cite{benhimane2004real} and so on. 
Recently, some methods \cite{moo2018learning,zhang2019learning,ma2019lmr} cast the task of finding correct correspondences as a binary classification problem and solve it by neural network. Specifically, these methods first obtain a putative correspondence set of an image pair by extracting local features and matching. Then the network takes the putative set as input and divides them into inliers (positive class) and outliers (negative class) and estimates relative pose , i.e., essential matrix ($E$ matrix) \cite{Hartley2004Multiple}. 

Due to various reasons (e.g., wide-baseline and illumination/scale changes), the number of outliers in the putative correspondence set is much larger than inliers, which usually results in a class imbalance problem of binary classification.
%In this case, it usually occurs that the precision and recall are extremely unbalanced in the classification results, which affects the network performance.
As shown in Fig. \ref{introduction}, we train the same network (CN-Net \cite{moo2018learning} is used) with two commonly used loss functions, including cross entropy loss (CE-Loss) and instance balance cross entropy loss (IB-CE-Loss) \cite{deng2018pixellink} on a class imbalance dataset, and present the training curves of precision and recall. 
These two loss functions can alleviate the class imbalance problem to some extent in some tasks, such as object classification \cite{he2016deep} and segmentation \cite{chen2017deeplab}. However, they lack a direct connection with precision and recall, making the network unable to dynamically adjust the bias of precision and recall. Fig. \ref{introduction} has demonstrated that, even though the precision and recall have been severely unbalanced during training, the loss functions can not adjust training direction to narrow the gap between precision and recall.
In fact, too low either precision or recall will lead to inaccurate relative pose estimation \cite{Hartley2004Multiple}. Thus, balanced precision and recall are very important.

The requirement to balance precision and recall can be transformed into a problem of maximizing Fn-measure, an evaluation criterion that considers both precision and recall.
In fact, Zhao et. al have already proposed to make Fn-measure differentiable and use it as loss function for salient object detection task \cite{zhao2019optimizing}. However, when replacing the IB-CE-loss of CN-Net \cite{moo2018learning} with Fn-measure, which has been verified in the subsequent experiments, the network gets a performance degradation. The degradation may be caused by the following two reasons: 1) Some relaxation is necessary to make Fn-measure differentiable to be loss function, which may cause the training becomes a sub-optimization process. 2) The network cannot make use of all the samples, because the $TN$ (true negative) samples are not related with the computation of Fn-measure. In other words, directly using Fn-measure as the loss function may abandon the advantages of cross-entropy loss.

In order to retain the advantage of cross entropy loss while maximizing Fn-measure, 
we propose a new Guided Loss which keeps the form of the cross entropy and use the Fn-measure as a guidance to adjust the optimization goals dynamically. We theoretically prove that a perfect negative correlation can be established between the loss and Fn-measure by dynamically adjusting the weights of positive and negative classes. Specifically, the perfect negative correlation is that the change in loss is completely opposite to the change in Fn-measure. Thus, with the decrease of the loss, the Fn-measure of the network will increase, so that the network is always trained towards the direction of increasing Fn-measure. By this way, the network maintains the advantage of the cross-entropy loss while maximizing the Fn-measure. It is worth mentioning when establishing the relationship between Fn-measure and loss, no relaxation is required, which is more advantageous than using Fn-measure as loss.

Besides loss function, another challenge is how to better encode global context in the network. Unlike 3D point clouds, not each correspondence contributes to the global context. In contrast, outliers are noises to the global context \cite{sun2020acne}. This issue is previously exploited by introducing spatial attention in the network \cite{Ploetz:2018:NNN,sun2020acne}. These methods learn a weight for each correspondence when encoding global context, so that the network can allow for outliers to be ignored. The key to these approaches is that the weight of outlier must be lower than inlier when encoding global context. However, learning appropriate weight for each correspondence in advance is a chicken-and-egg problem. In the shallow layers of the network, it is hard to learn appropriate weight because the features in these layers are less recognizable.
In fact, Lowe Ratio \cite{lowe2004distinctive}, i.e., the side information generated during feature matching, is proved to be powerful prior information to determine the confidence of each point being inlier \cite{goshen2008balanced,brahmachari2009blogs}. Based on this observation, we propose a Bayesian attentive context normalization (BACN) to mine prior information for better reducing the noise of outliers to global context. The prior can be integrated into the network to better encode global context. Besides, to capture more complex channel-wise context, we generalize the channel-wise attention (CA) \cite{hu2018squeeze} operation and reshape it as a point-wise form through group convolution \cite{cohen2016group}. The BACN and CA are further combined as a hybrid attention block for feature extraction.

Since the proposed Guided Loss can change the network's  bias toward precision and recall by using different Fn-measures (set $n$ as different value) as guidance, we can build a cascade network by the Guided Loss. Specifically, we first train the network through a Fn-measure with big $n$ as the guidance to obtain a coarse result with high recall. So the network keeps as many inliers as possible while filtering out some outliers. After that, Fn-measure with a smaller $n$ can be used as guidance to optimize the coarse result. As $n$ gets smaller, the network gradually leads to a result with higher precision. By gradually optimizing the result from coarse to fine, the network can achieve a better performance than that obtained by one fixed Fn-measure Guided Loss.

%Based on HA Block and Guided Loss, we propose a cascade network, which is an end-to-end network for two view geometry estimation. Specifically, the cascade network first gets a coarse result and then optimizes the coarse result step by step. Obviously, we would prefer to keep more inliers in the coarse result, while eliminating some significant outliers. In the final result, the network optimizes the coarse result to obtain a  high-precision correspondence set. since Guided Loss can change the network's  bias toward precision and recall through using different Fn-measures as guidance, we can use different Fn-measures Guided Loss to supervise coarse  and fine results. For the coarse result, we use Fn-measure with bigger n as the guidance to remain more inliers. For the fine result, a Fn-measure with smaller n is utilized as the guidance to obtain a result with high precision. Thus, the cascade network can better find the good correspondences. In a nutshell, our contributions is threefold: 
In a nutshell, our contribution is threefold: (i) We propose a novel Guided Loss for two-view geometry network. It can establish a direct connection between loss and Fn-measure, so the network can better optimize Fn-measure. (ii) We design a hybrid attention block to better extract global context. It combines a Bayesian attentive context normalization and a channel-wise attention to capture the low-level prior information and channel-wise awareness. 
(iii) Based on the Guided Loss and hybrid attention block, we design a cascade network for two-view geometry estimation. Experiments  show that our network achieves state-of-the-art performance on benchmark datasets.

\section{Related Works}

\textbf{Model fitting methods} usually determine inliers by judging whether the raw matches satisfy the fitted epipolar geometric model. The classic RANSAC \cite{fischler1981random} adopts a hypothesize-and-verify pipeline, so do its variants, such as PROSAC \cite{chum2005matching}. Besides, many modifications of RANSAC have been proposed. Some methods \cite{chum2005matching,fragoso2013evsac,brahmachari2009blogs,goshen2008balanced} mine prior information to accelerate convergence. Some other methods \cite{chum2003locally,barath2018graph} augment the RANSAC by performing a local optimization step on the so-far-the-best model.

% However, these methods can not deal with the data with the low ratio of inliers. What's more, some complex models cannot be expressed for a single epipolar geometric model, such as multi-consistency matching \cite{xiao2019superpixel}.

%\noindent \textbf{Non-parametric Methods} 

\noindent\textbf{Learning Based Methods.}
Since deep learning has been successfully applied for dealing with unordered data \cite{qi2017pointnet,qi2017pointnet++}, learning based methods attract great interest in two-view geometry estimation.
CN-Net \cite{moo2018learning} reformulates the mismatch removal task as a binary classification problem. It utilizes a simple Context Normalization (CN) operation to extract global context. Based on CN, some network variants are proposed. NM-Net \cite{zhao2019nm} employs a simple graph architecture with an affine compatibility-specific neighbor mining approach to mine local context. $ \rm N^{3}$-Net \cite{Ploetz:2018:NNN} presents a continuous deterministic relaxtaion of KNN selection and a $\rm N^{3}$ block to mine non-local context. OA-Net \cite{zhang2019learning} utilizes an Order-Aware network to build model relation between different nodes. ACN-Net \cite{sun2020acne} introduces spatial attention to two-view geometry network. 
Our work is to mine prior information and channel-wise awareness to improve the performance of the network.

%In this paper, we further develop the prior information mining and channel awareness in our hybrid attention block.  

\noindent\textbf{Attention Mechanism} focuses on perceiving salient areas similar to human visual systems \cite{vaswani2017attention}. Non-local neural network \cite{wang2018non} adopts non-local operation to introduce attention mechanism in feature map. SE-Net \cite{hu2018squeeze} introduces channel-wise attention mechanism through a Squeeze-and-Excitation block. In order to explore second-order statistics, SAN-Net \cite{dai2019second} utilizes second-order channel attention (SOCA) operations in their network. In addition to the two dimensional convolution, Wang et. al propose a graph attention convolution (GAC) \cite{wang2019graph} for dealing with point cloud data. 
%The above literature shows that attention mechanism can enhance the network performance of different tasks. In this paper, we are committed to design a suitable attention module for two-view geometry estimation.

\section{Method}
\subsection{Problem Formulation}
Given an image pair, we first extract local features (handcrafted descriptors such as SIFT \cite{lowe2004distinctive}, or deep learning based descriptors, such as Hard-Net \cite{mishchuk2017working}) of each image and perform feature matching to establish a set of putative correspondences between them. The coordinates of each correspondence in the putative set are concatenated as the input of our network, as follows: 
\begin{equation}\label{input}
	C=[c_{1};c_{2};...,c_{N}] \in \mathbb{R}^{N \times 4}, c_{i}=(x_{1}^{i}, y_{1}^{i}, x_{2}^{i}, y_{2}^{i}),
\end{equation}
where $N$ is the number of putative correspondences. $(x_{1}^{i}, y_{1}^{i})$ and $(x_{2}^{i}, y_{2}^{i})$ are the coordinates of the two feature points of $i$-th correspondence. The coordinate of each feature point is normalized by camera intrinsics \cite{moo2018learning}. 
The network extracts a feature for each correspondence and determines the probability that a correspondence is inlier based on their features as follows: 
\begin{equation}\label{network}
	L = \Phi(C), L \in {\mathbb{R}}^{N \times 1},
\end{equation}
where $\Phi(\cdot)$ is the network with trained parameters. $L$ is the logit values predicted by the network.
After that, the network performs a differentiable weighted eight-point algorithm \cite{moo2018learning} on the correspondence to estimate the relative pose ($E$ matrix). as follows:
\begin{equation}\label{network}
	\begin{aligned}
		\hat{E} = g(w, C),  w= &\tanh({\rm ReLU}(L)),
	\end{aligned} 
\end{equation}
$g(\cdot, \cdot)$ is the weighted eight-point algorithm, $L$ is the predicted logit value and $\hat{E}$ is the estimated $E$ matrix.

\subsection{Guided Loss}
\label{headings}
The correspondence classification in our network is a binary classification task. In general, the result is evaluated by the Fn-measure ($Fn$), which considers both precision ($P$) and recall ($R$), as follows:
\begin{equation}\label{Fn}
	Fn = {(1 + n^{2}) \cdot P \cdot R}/(n^{2} \cdot P + R).
\end{equation}
When $n > 1$, the Fn-measure is biased in favour of recall and otherwise in favour of precision.
When adopting cross entropy loss as objective function, the loss will gradually decrease under the successive optimization. However, there is no guarantee that a drop in the loss will result in an increase of Fn-measure. Therefore, the network may not be trained towards the direction of optimizing Fn-measure. Based on this observation, we propose a hypothesis, that is, whether the relationship between the cross entropy loss and Fn-measure can be established, so that the decrease of loss will lead to the increase of Fn-measure.
This relationship can be expressed in the form of differential as follows:
\begin{equation}\label{dldf}
	dl \cdot dFn \leq  0.
\end{equation}
Specifically, the proposed Guided Loss ($l$) uses the form of IB-CE-loss as follows:
\begin{equation}\label{rewritten_loss}
	\begin{aligned}
		l=-(\lambda \frac{1}{N_{pos}} \sum_{i=1}^{N_{pos}}\log(y_{i}) + & \mu  \frac{1}{N_{neg}} \sum_{j=1}^{N_{neg}}\log(1-y_{j})), \\
		s.t. \quad \ \lambda + \mu = 1, & N_{pos} + N_{neg} = N 
	\end{aligned}
\end{equation}
where $N_{pos}$ and $N_{neg}$ are the number of positive and negative samples. $\lambda$ and $\mu$ are the weights of positive and negative samples. $y_{i}$ and $y_{j}$ is the logit value of correspondence $i$ and $j$ respectively.
Meanwhile, after forward propagation of the network, all the samples are divided into four categories, including $FP$ (false positive), $FN$ (false negative), $TP$ (true positive) and $TN$ (true negative). Suppose the number of $FN$ and $FP$ samples are $X$, $Y$ respectively, then the number of $TP$ and $TN$ can be computed as follows:
\begin{equation}\label{TPTN}
	N_{TP} = {N_{pos} -X}, N_{TN} = {N_{neg} -Y},
\end{equation}
and the precision ($P$) and recall ($R$) in Fn-measure ($P$, $R$ in Eq. \ref{Fn}) can be computed as follows:
\begin{equation}\label{PR}
	\begin{aligned}
		P = ({N_{pos} -X})&/(N_{pos} - X + Y), \\ 
		R = (N_{pos} &- X) /{N_{pos}},
	\end{aligned}
\end{equation}
Thus, Fn-measure is the dependent variable of $X$ and $Y$ according to Eq. \ref{Fn} and \ref{PR}. We express the functional relationship between Fn-measure ($Fn$) and $X, Y$ as follows:
\begin{equation}\label{F_X_Y}
	Fn = {\rm {F}}(X, Y).
\end{equation}
In order to derive the relationship between Fn-measure and the loss, we also expect to express the loss as the dependent variables of $X$ and $Y$. In the forward propagation of the network, we can calculate the average loss terms of $TP$, $TN$, $FP$ and $FN$ samples respectively, denoted as $l_{TP}, l_{TN}, l_{FP}, l_{FN}$. Then the loss in Eq. \ref{rewritten_loss} can be equivalently calculated as follows:
\begin{equation}\label{simplify_loss}
	\begin{aligned}
		l = \lambda/N_{pos} \cdot \{X \cdot l_{FN} +
		(N_{pos} - X) \cdot l_{TP}\} \\
		+ \mu/N_{neg} \cdot \{Y \cdot l_{FP} + (N_{neg} - Y) \cdot l_{TN}\}
	\end{aligned}  
\end{equation}
We compute the derivative forms of loss function ($dl$) and Fn-measure ($dFn$) by $X$ and $Y$ as follows:
\begin{equation}\label{detail}
	\begin{aligned}
		dl &= \partial l_{X} dX + \partial l_{Y} dY, \\
		dFn &= \partial F_{X} dX + \partial F_{Y} dY,
	\end{aligned}
\end{equation}
where $\partial l_{X}$ and $\partial l_{Y}$ are the partial derivatives of loss with respect to $X$ and $Y$, and $\partial F_{X}$ and $\partial F_{Y}$ are the partial derivatives of Fn-measure with respect to $X$ and $Y$.
Then, we can draw a sufficient condition of Eq. \ref{dldf} as follows: 
\begin{equation}\label{st}
	\partial F_{X}/{\partial F_{Y}} = \partial l_{X}/{\partial l_{Y}}.
\end{equation}
\textbf{Algorithm.} 
The $\partial l_{X}$ and $\partial l_{Y}$ can be computed according to Eq. \ref{simplify_loss} as follows:
\begin{equation}\label{p_l_x}
	\begin{aligned}
		\partial l_{X} = \lambda/N_{pos} \cdot (l_{FN} - l_{TP}),  \\
		\partial l_{Y} = \mu/N_{neg}\cdot (l_{FP} - l_{TN}).
	\end{aligned}  
\end{equation}
Meanwhile, $\partial F_{X}$ and $\partial F_{Y}$ can also be calculated by means of numerical derivatives (step 4 in Algorithm \ref{alg:G_Loss} ) in the training process. Obviously, to hold Eq. \ref{st}, the weights $\lambda$ and $\mu$ should be dynamically changed during training. This also reveals the problem of IB-CE-Loss using a fixed $\lambda$ and $\mu$ during training. In order to establish a relationship between loss and Fn-measure as Eq. \ref{dldf}, we design a weight adjustment algorithm by making Eq. \ref{st} hold, as follows:
\begin{algorithm} [h]%算法开始 
	\caption{Guided Loss} %算法的题目 
	\label{alg:G_Loss} %算法的标签 
	%	\algsetup{linenosize=\footnotesize}
	%	\footnotesize
	{\bf Input:} The classification result after forward propagation\\
	{\bf Output:} Weights of positive and negative samples in loss function ($\lambda$ and $\mu$)  
	\begin{algorithmic}[1] %此处的[1]控制一下算法中的每句前面都有标号 
		\FOR{$i = 0; i < Batch\_size; i ++$}
		\STATE Count the number of $N_{pos_{i}}$ and $N_{neg_{i}}$. Count the number $TP$, $TN$, $FP$, $TN$ samples as $N_{TP_{i}}$, $N_{FP_{i}}$, $N_{TN_{i}}$, $N_{FN_{i}}$, then $X_{i} = N_{FN_{i}}$, $Y_{i}=N_{FP_{i}}$.

		\STATE Compute the average loss of $TP$, $TN$, $FP$ and $FN$ samples as $l_{TP_{i}}$, $l_{TN_{i}}$, $l_{FP_{i}}$ and $l_{FN_{i}}$.
		
		\STATE Compute $\partial F_{X_{i}}$ and $\partial F_{Y_{i}}$: $\partial F_{X_{i}}$ = ${\rm {F}}(X_{i} + 1, Y_{i}) - {\rm {F}}(X_{i}, Y_{i})$, $\partial F_{Y_{i}}$ = ${\rm {F}}(X_{i}, Y_{i} + 1) - {\rm {F}}(X_{i}, Y_{i})$
		
		\STATE s.t. $\lambda_{i} + \mu_{i} = 1$ $\rightarrow$ compute $\lambda_{i}$ and $\mu_{i}$ according to Eq. \ref{st}, \ref{p_l_x} and step 2, 3 and 4
		\ENDFOR
		\STATE return $\lambda$, $\mu$
	\end{algorithmic} 
\end{algorithm}

Specifically, when a batch of training data is sent to the network, the first step is forward propagation. After the forward propagation, we can use algorithm \ref{alg:G_Loss} to get $\lambda$ and $\mu$ for making Eq. \ref{st} hold. Then we substitute $\lambda$ and $\mu$ into Eq. \ref{rewritten_loss} and perform back propagation.

\begin{figure*}[t]
	\centering
	\includegraphics[width=2\columnwidth]{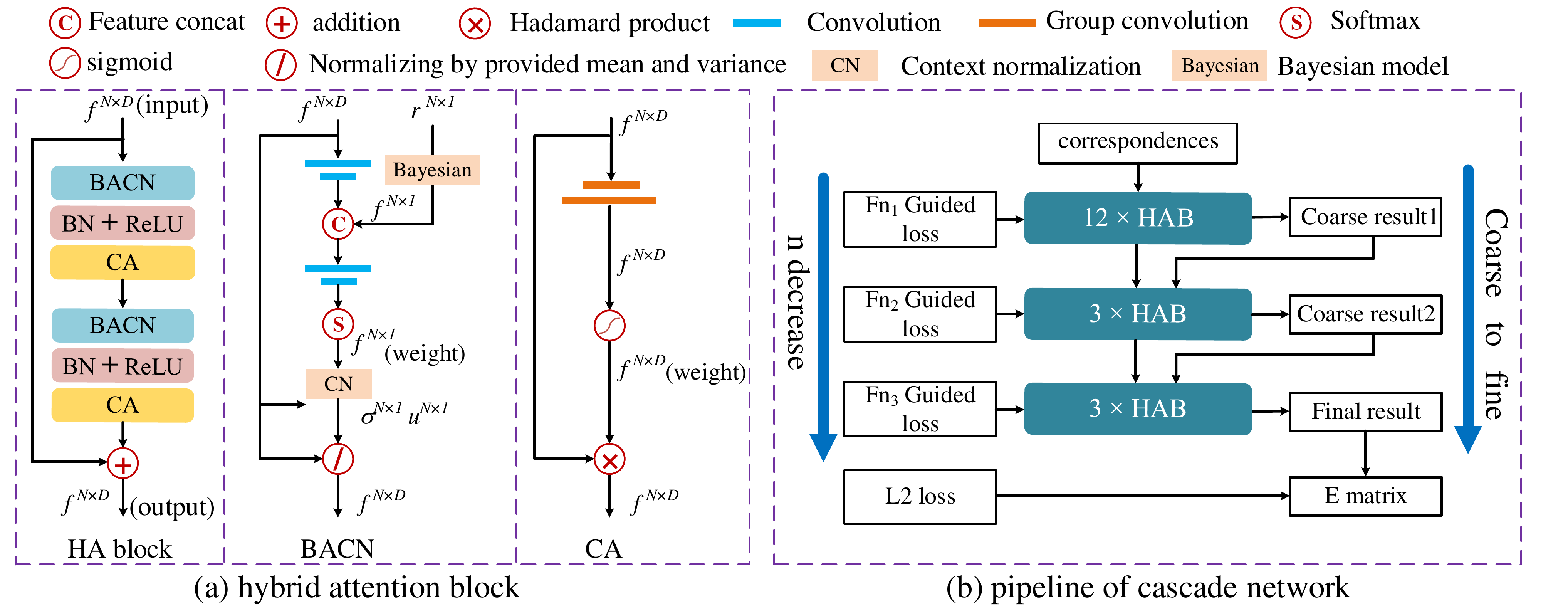} % Reduce the figure size so that it is slightly narrower than the column. Don't use precise values for figure width.This setup will avoid overfull boxes. 
	\caption{\textbf{Network architecture}. (a) The hybrid attention block (HAB) is made up of Bayesian attentive context normalization (BACN), batch normalization, ReLU and channel-wise attention (CA) in a Res-Net \cite{he2016deep} architecture. (b) The pipeline of the cascade network.} 
	\label{pipeline}
\end{figure*}

\subsection{Hybrid Attention Block}
The basic feature extraction block of our network is the proposed hybrid attention block (HAB). As shown in Fig. \ref{pipeline} (a), the input of the HAB is the feature map $f^{N \times C}$ (output of last layer or data points at layer zero), where $N$ is the number of correspondences and $C$ is the number of channels. HAB integrates Bayesian attentive context normalization (BACN), batch normalization (BN) \cite{ioffe2015batch}, ReLU and Channel-wise Attention (CA) operations in the structure of Res-Net \cite{he2016deep}. Specifically, BACN is to normalize each correspondence so that the features of correct and incorrect correspondences are distinguishable. BN is adopted to accelerate network convergence and ReLU function is utilized as an activation function. Finally, the CA operation learns the statistical information on the channel to boost the performance of the network.

\noindent\textbf{Bayesian Attentive Context Normalization.} 
We first briefly introduce how our BACN learns distinguishable features for inliers and outliers. In fact, the inliers are under the constraint of an $E$ matrix while outliers are not \cite{Hartley2004Multiple}. 
In BACN, a global context is utilized to replace the constraint of $E$ matrix to normalize each correspondence. 
%Specifically, as shown in Fig. \ref{pipeline} (a), we consider the weighted mean and variance of all the features as global context, and use the context normalization \cite{moo2018learning} operation to encord feature for each correspondence, as follows:
We use statistical information, i.e., the mean and variance of all features, as the global context. Since the global context is expected to fit the distribution of inliers, we use a weighted mean and variance as the global context, so that the outliers can be ignored by the weight vector. Then, we use the context normalization \cite{moo2018learning} operation to encode feature for each correspondence, as follows: 
\begin{equation}\label{BACN}
	{\rm CN}(f_{i}^{l}) = {(f_{i}^{l} - u^{l})} / {\sigma^{l}},
\end{equation}
where $f_{i}^{l} \in \mathbb{R}^{C}$ is the feature of correspondence $i$ in $l$-th layer. $u^{l}$ and $\sigma^{l}$ are the weighted mean and variance.

The key to BACN is how to better learn the weight vector for computing weighted mean and variance. In the shallow layers of the network, it's hard to learn appropriate weight vector because the features in these layers are less recognizable. Since Lowe Ratio \cite{lowe2004distinctive}, which is generated during feature matching, is proved useful for determining the confidence of each correspondence being inlier, we expect to use it to make up for the dilemma of weight learning in shallow network.
%The independence and identical distribution of features is a very important assumption in neural networks \cite{Bishop2006PRML}. 
However, the distribution of Lowe Ratio is quite different on different datasets, while independence and identical distribution of features is a very important assumption in neural networks \cite{Bishop2006PRML}. To make better use of Lowe Ratio, we first use the Bayesian Model \cite{Bishop2006PRML} to convert the it into a probability value. Formally, given a pair of correspondence with Lowe Ratio $r_{i} \in \mathbb{R}^{1}$, we consider $r_{i}$ as a variable and the joint probability distribution function (PDF) can be modeled as:
\begin{equation}
	\begin{aligned}\label{Prior}
		f_{r}(r_{i})=f_{in}(r_{i})\alpha+f_{out}(r_{i})(1-\alpha),
	\end{aligned}
\end{equation}
where $f_{in}(r_{i})=f(r_{i}| r_{i}$ belongs to an inlier$)$, $f_{out}(r_{i})=f(r_{i}|r_{i}$ belongs to an outlier$)$, and $\alpha$ is the inlier ratio of the putative correspondence set of a specific image pair. 
Then, the prior probability $p_i(in)$ that the $i$-th correspondence belongs to inlier can be calculated as follows:
\begin{equation}
	\begin{aligned}\label{Prior2}
		p_i(in)=f_{in}(r_{i})\alpha / \{f_{in}(r_{i})\alpha + f_{out}(r_{i})(1-\alpha)\}.
	\end{aligned}
\end{equation}
Before training, we obtain the PDF of inlier ($f_{in}$) and outlier ($f_{out}$) on the training dataset with ground-truth as empirically PDF. Then for each image pair, we estimate the inlier ratio
$\alpha$ using a curve fitting method \cite{goshen2008balanced}. Thus we assign a prior probability to each correspondence by Eq. \ref{Prior2}. 

After obtaining the prior probability for each correspondence, it will be utilized to participate in the calculation of weight vector. The architecture of weight learning is inspired by Bayesian Model \cite{Bishop2006PRML}. The prior probability is similar to the prior probability of Bayesian Model. Meanwhile, as shown in Fig. \ref{pipeline} (a), the input of BACN is followed with two convolution operation to learn a temporary weight vector, which is similar to the likelihood probability of Bayesian Model. The prior probability and the likelihood probability are fused through a feature concatenate operation to generate a feature which encode the information of posterior probability. After that, the posterior feature is followed by two convolution and a softmax operations to produce weight vector. By
incorporating prior information, our network is easier to obtain better classification results.

\noindent\textbf{Channel-wise Attention.} The statistics on the channel have been shown to have a significant impact on the network \cite{hu2018squeeze,wang2019graph}. In order to enhance the channel awareness of the network, we introduce channel-wise attention to the HA block. We learn a channel weight vector for each correspondence instead of a weight vector that is shared by all the correspondences to capture complex channel context. When learning the weight vector, group convolution \cite{cohen2016group} is used to reduce network computation. Formally, Let $f_{i}^{l} \in \mathbb{R}^{C}$ be the feature of correspondence $i$ in $l$-th layer, then the CA can be expressed as follows: 
\begin{equation}
	{\rm CA} (f_{i}^{l}) = f_{i}^{l} * w_{i}, i = 1,...N,   \\
\end{equation}
where $w_{i}$ is obtained by performing two group convolution operations \cite{cohen2016group} and a sigmoid function on the feature as shown in Fig. \ref{pipeline} (a).

%\textbf{Hybrid Attention Block.} The BACN and CA are combined in Res-Net architecture as the feature extraction block, called hybrid attention block (HA Block) as Fig. \ref{pipeline} (a) . It is utilized as the basic feature extraction block in our network.

\subsection{Cascade Architecture} 
%Introducing supervisory information in the middle layer of the network is conducive to network convergence \cite{simonyan2014very}. However, when adopting cross entropy loss as auxiliary loss in the middle layer, the network, it is likely to result in a low recall and many inliers are discarded after this layer. 

%Since the proposed Guided Loss can flexibly control the bias on precision and recall by using different Fn-measure as guidance, we adopting Guided Loss as auxiliary loss to build a cascade network. In the middle layer, we use F3-measure Guided Loss, which focus more on recall, as auxiliary loss o better learn the parameters of the shallow network.

Since the proposed Guided Loss can flexibly control the bias on precision and recall by using different Fn-measure as guidance, we can naturally build a cascade network  by Guided Loss to progressively refine the performance. 
Specifically, as shown in Fig. \ref{pipeline} (b), we first use a 12-layer hybrid attention blocks as feature extraction module to extract the feature for each correspondence. Then a coarse result (coarse result1 in Fig. \ref{pipeline} (b)) can be obtained through these features by $\rm{Fn_{1}}$-measure Guided Loss. 
Then two refinement modules are followed to perform local optimization to refine the coarse result. 
Each refinement module is made up of a 3-layer HA Block. Different from feature extraction module, the global context in refinement module is extracted from the coarse result of the previous module instead of all of the correspondences. Besides, in order to gradually optimize the coarse result, the loss function will also progressively bias the precision. The coarse result2 is obtained by $\rm{Fn_{2}}$-measure Guided Loss, and the final result is obtained by $\rm{Fn_{3}}$-measure Guided Loss. During training, $n_{1} > n_{2} > n_{3}$ holds so that the network gradually obtains result with higher precision. Finally, the $E$ matrix is computed by performing weighted eight-point or RANSAC algorithm on the final result, and it is supervised by a $L2$-$loss$.

\begin{table*}
	\caption{Comparison with other baselines on YFCC100M$\&$SUN3D and COLMAP dataset. \textbf{P}recision (\%), \textbf{R}ecall (\%), \textbf{F1}-measure (\%) and mAP (\%) under $5^{\circ}$ and $10^{\circ}$  (\textbf{with weighted eight-point algorithm/with RANSAC}) are reported.}
	\label{tab:overall_performance}
	%	\footnotesize
	\centering
	\resizebox{0.99\linewidth}{!}
	{
		\begin{tabular}{c|ccccc||ccccc}
			\hline 
			& \multicolumn{5}{c||}{YFCC100M$\&$SUN3D}  & \multicolumn{5}{c}{COLMAP}  \\ 
			
			\cline{2-11}  
			& P & R & F1 & mAP $5^{\circ} $ & mAP $10^{\circ} $ & P & R & F1 & mAP $5^{\circ} $ & mAP $10^{\circ} $  \\
			
			\hline
			
			%Ransac & 23.74 & 40.83 & 29.32 & -/4.00 & -/6.82 & 25.16 & 14.48 & 18.72 & -/2.28 & -/4.52 \\
			CN-Net & 37.23 & 73.21 & 47.08 & 15.12/33.11 & 31.87/43.47 & 27.95 & 65.63 & 39.82 & 11.82/26.89 & 18.44/30.82  \\ 
			PointNet & 27.83 & 47.23 & 32.48 & 12.12/26.31 & 27.85/33.92 & 13.60 & 41.77 & 27.66 & 10.41/25.65 & 17.94/28.76 \\
			ACN-Net & 41.09 & \textbf{80.96} & 50.68 & 27.91/37.18 & 37.86/47.54 & 30.13 & \textbf{76.83} & 38.10 & 23.15/32.51 & 27.32/35.88 \\ 
			%\checkmark & \checkmark & \checkmark & & & 28.2/38.5 \\ \hline
			NM-Net & 40.66 & 71.66 & 50.74 & 17.70/34.09 & 32.80/42.92 & 31.99 & 54.48 & 38.92 & 20.96/31.72 & 23.08/33.42 \\ 
			$\rm N^{3}$-Net & 40.92 & 75.34 & 51.68 & 14.52/32.65 & 30.27/42.16 & 26.88 & 62.91 & 33.26 & 10.90/25.68 & 16.74/29.77
			\\ 
			
			OA-Net & 40.88 & 72.33 & 48.58 & 30.53/37.80 & 39.84/49.87 & 37.41 & 57.74 & 42.88 & 26.82/34.57 & 29.99/37.09 
			\\ 
			
			Ours & \textbf{53.46} & 70.59 & \textbf{59.67} & \textbf{31.25/41.90} & \textbf{41.52/52.57} & \textbf{42.08} & 55.21 & \textbf{46.76} & \textbf{27.82/36.83} & \textbf{30.80/39.26} \\
			
			\hline
		\end{tabular}
	}
\end{table*}

\begin{figure*}
	\centering
	%	\subfigure[weight $\lambda$ ]{\includegraphics[width=2.7in]{fig//trainingPR4_weight3.pdf}}
	%	\subfigure[Precision and recall]{\includegraphics[width=2.7in]{fig//trainingPR4_3.pdf}}
	\subfigure[weight $\lambda$]{\includegraphics[width=0.47\textwidth]{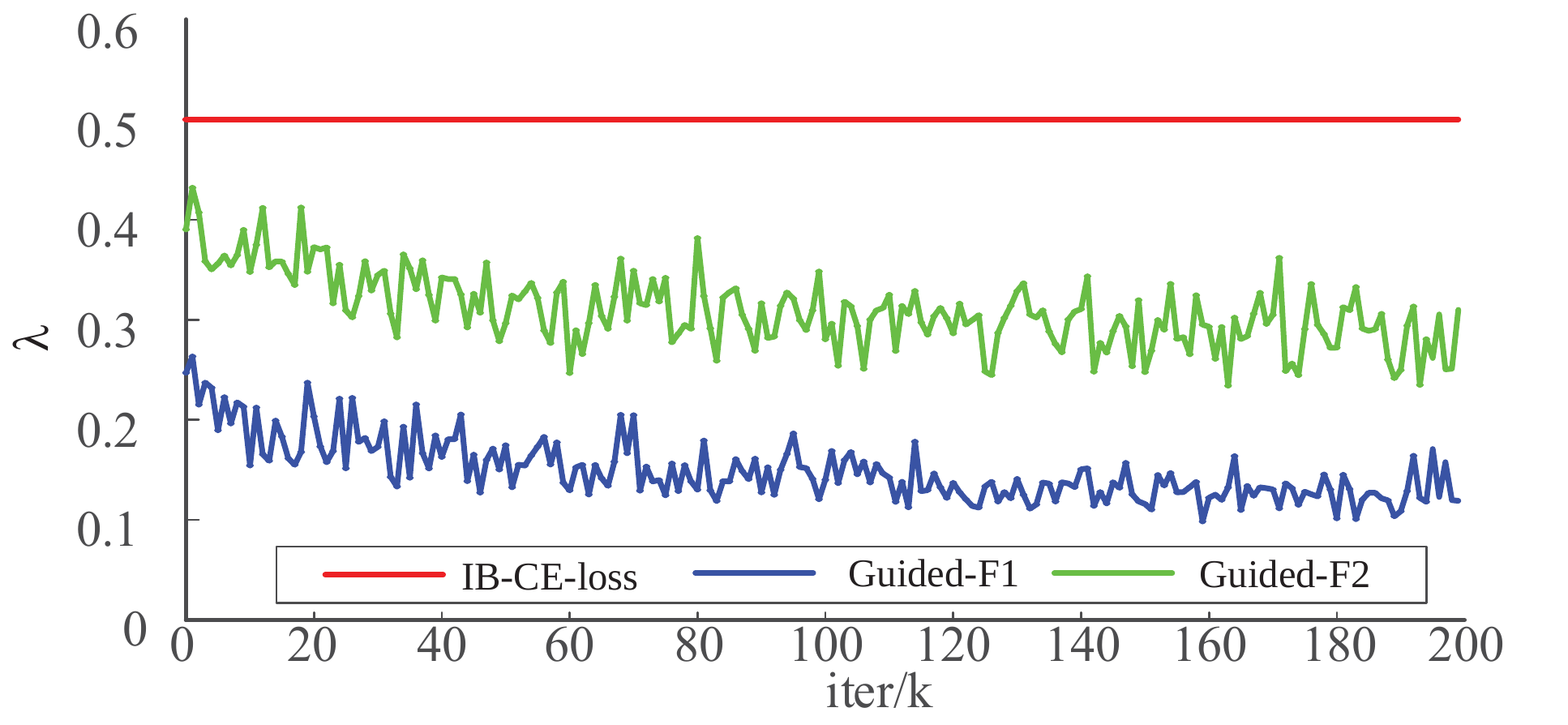}} 
	\subfigure[Precision and recall]{\includegraphics[width=0.47\textwidth]{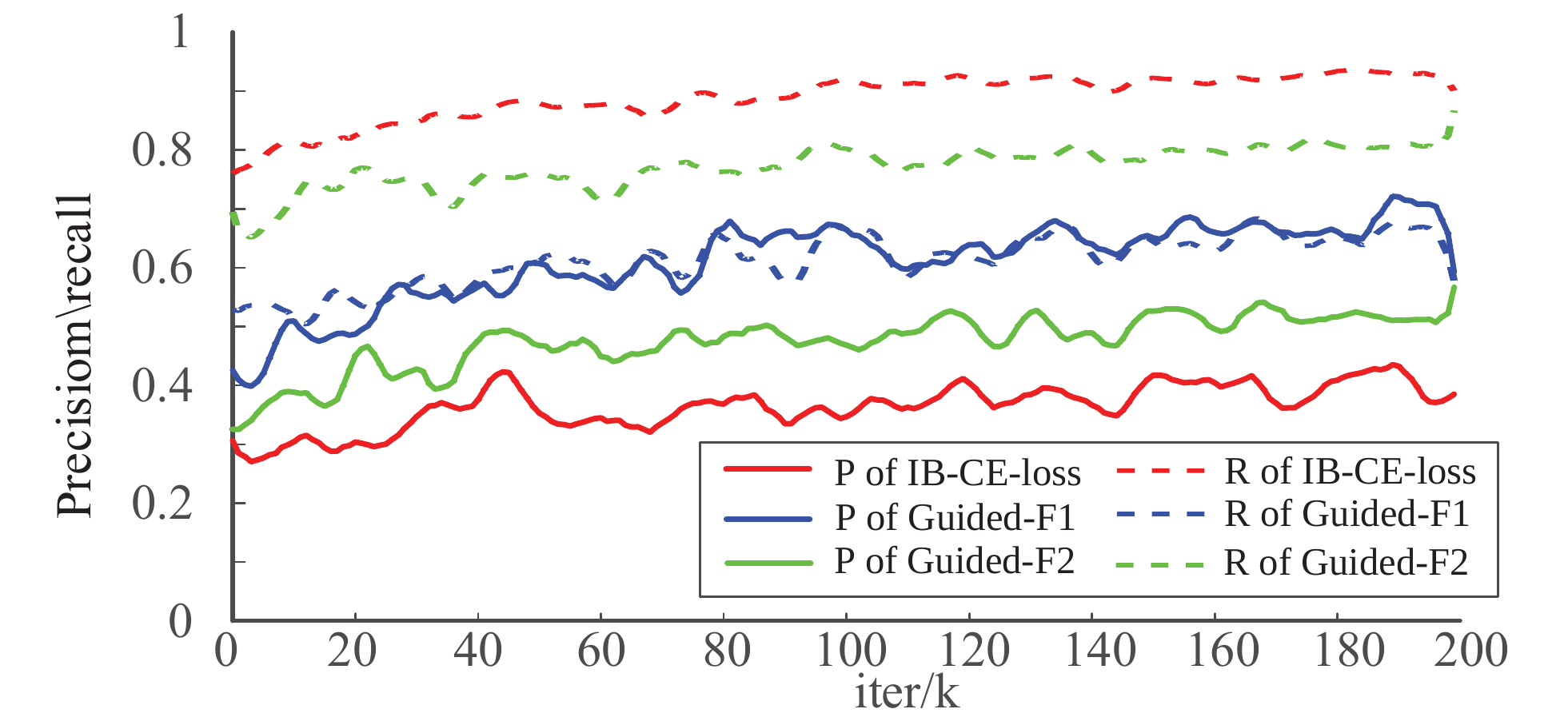}}
	\caption{\textbf{Training curves.} We train the baseline CN-Net \cite{moo2018learning} with different classification loss on YFCC100M$\&$SUN3D dataset. (a) The weight curve of positive class ($\lambda$ in Eq. \ref{rewritten_loss}). (b) The precision and recall curves on the  validation set. We only record the curve of $\lambda$ because the sum of $\lambda$ and $\mu$ is always 1 during training.}
	\label{training_curve}
\end{figure*}

\noindent\textbf{Loss Function.} We formulate our objective as a combination of two types of loss functions, including classification and
regression loss. The whole loss function is as follows:
\begin{equation}
	\begin{aligned}\label{wholeLoss}
		loss=l_{cls}+\eta_{1} l_{cls1}+\eta_{1} l_{cls2}+\eta_{3} l_{reg}.
	\end{aligned}
\end{equation} 
$l_{cls}$ is related with the final result in Fig. \ref{pipeline}, and $l_{cls1}$ and $l_{cls2}$ are related with the coarse result1 and coarse result2 in Fig. \ref{pipeline} respectively. For the regression loss $l_{reg}$, we use geometric $L2$-$loss$ for $E$ matrix \cite{moo2018learning} as follows:
\begin{equation}
	\label{loss_reg}
	l_{reg}=min\{\left \| \hat{E} \pm E \right \| \}, 
\end{equation} 
where $\hat{E}$ and $E$ are the estimated and ground truth $E$ matrix, respectively.

\section{Experiments}
%In this section, we conduct extensive experiments for both correspondence classification and E matrix regression on benchmark datasets. All experiments are run on Ubuntu 16.04 with NVIDIA GTX1080Ti. 

\subsection{Experimental Setup}
\textbf{Parameter Settings.} 
The network is trained by Adam optimizer \cite{kingma2014adam} with a learning rate being $10^{-3}$ and batch size being 16. The iteration times are set to 200k. In Eq. \ref{wholeLoss}, the loss weight $\eta_{3}$ is 0 during the first 20k iteration and then 0.1 later. $\eta_{1}$ and $\eta_{2}$ are set to 0.1 during the whole training. The $n_{1}$, $n_{2}$ and $n_{3}$ in Fig. \ref{pipeline} are set to 0.3, 0.25 and 0.2 during training, which leads to best relative pose estimation results.

% We only use BACN in the first 4 layers, because moderating a prior information helps improve performance, but too much can degrade \cite{goshen2008balanced}.

\noindent\textbf{Datasets.} 
We mainly evaluate our method on two benchmark datasets. The first dataset is YFCC100M$\&$SUN3D dataset \cite{moo2018learning}. Yi et al. choose 5 scenes from the YFCC100M dataset \cite{thomee2016yfcc100m} as outdoor scene and 16 scenes from SUN3D dataset \cite{xiao2013sun3d} as indoor scene. The ground truth for outdoor and indoor scenes is generated from VSfM \cite{wu2013towards} and KinectFusion \cite{newcombe2011kinectfusion}. We use their dataset and exact data splits. 
The second dataset is COLMAP dataset, which is published by Zhao et. al \cite{zhao2019nm}.
It contains 16 outdoor scenes. We also use their dataset and exact data splits. 

\noindent\textbf{Evaluation Criteria.} 
In the test, we use the trained classification network to get the correspondence classification results, and employ the precision (P), recall (R), F1-measure (F1) \cite{van1974foundation} as the evaluation criteria. Then we perform weighted eight-point \cite{moo2018learning} and RANSAC \cite{fischler1981random} methods as post-processing on the classified correspondence to recover the $E$ matrix between the image pair. In order to evaluate the results of the relative pose estimation, we recover the rotation and translation vectors from the estimated $E$ matrix and report mAP under $5^{\circ}$, $10^{\circ}$ as the metrics respectively \cite{moo2018learning,zhang2019learning}.

\begin{table*}[t]%%If you do not follow these requirements, your paper will be subject to expensive reformatting and special handling fees that can easily exceed the extra page fee.
	\centering%
	\caption{Ablation study on YFCC100M$\&$SUN3D datasets. \textbf{P}recision (\%), \textbf{R}ecall (\%), \textbf{F1}-measure (\%) and mAP (\%) under $5^{\circ}$ and $10^{\circ}$ (using \textbf{weighted eight-point/RANSAC} as post-processing) are reported.  \textbf{BACN}: Bayesian attention context normalization. \textbf{CA}: channel-wise attention. \textbf{ACN}: attentive context normalization \cite{sun2020acne}. \textbf{G-Loss}: Guided Loss. \textbf{No cas}: 18-layer network without cascade architecture. \textbf{Cas:} 18-layer network with cascade architecture.} \label{tab:Ablation}%
	
	\begin{tabular}{ccccccc|ccccc}
		\hline 
		\multicolumn{7}{c|}{module} & \multicolumn{5}{c}{result}  \\ 
		\cline{1-7} \cline{8-12}
		CN-Net & BACN & CA & ACN &  G-Loss & No-cas & cas & P & R & F1 & mAP $5^{\circ} (\%)$ & mAP $10^{\circ} (\%)$ \\ \hline 
		\checkmark & & & & & & & 37.23 & 73.21 & 47.08 & 15.12/33.11 
		& 31.87/43.47 \\ 
		\checkmark & \checkmark &  & & & & & 43.06 & 79.89 & 52.63 & 26.32/36.14 & 36.41/46.88 \\ 
		
		\checkmark & & \checkmark & & & & & 42.16 & 81.22 & 52.76 & 26.91/36.85 & 37.02/47.48 \\ 
		\checkmark & \checkmark & \checkmark & & & & & 44.08 & 81.35 & 53.48 & 27.83/37.99 & 37.82/48.29 \\
		
		\checkmark & & & \checkmark & & & & 40.20 & 80.01 & 50.48 & 25.87/35.68 & 35.66/46.04 \\ 
		\checkmark & \checkmark & \checkmark & & \checkmark & & & 51.32 & 68.27 & 57.55 & 29.95/39.88 & 39.38/50.02 \\ 
		
		\checkmark & \checkmark & \checkmark & & \checkmark & \checkmark & & 52.09 & 68.20 & 57.92  & 30.32/40.18 & 40.25/50.51\\
		
		%\checkmark & \checkmark & \checkmark & & & 28.2/38.5 \\ \hline
		\checkmark & \checkmark & \checkmark & & \checkmark & & \checkmark & {53.46} & 70.59 & 59.67 & {31.25/41.90} & 41.52/52.57 \\
		\hline 
	\end{tabular}

\end{table*}%\usepackage{times}

\subsection{Comparison to Other Baselines}

We compare our network with other state-of-the-art methods, including CN-Net \cite{moo2018learning}, PointNet \cite{qi2017pointnet}, ACN-Net \cite{sun2020acne}, NM-Net \cite{zhao2019nm}, $\rm N^{3}$-Net  \cite{Ploetz:2018:NNN} and OA-Net \cite{zhang2019learning} on both YFCC100M$\&$SUN3D and COLMAP datasets. All the networks are trained with the same setting. Tab. \ref{tab:overall_performance} summarizes the correspondence classification and relative pose estimation results. Our method shows improvement of 15.59\% and 6.94\% over CN-Net (baseline of our network) in terms of F1-measure on both YFCC100M$\&$SUN3D and COLMAP datasets. In terms of pose estimation results, the mAP of our network is also better than CN-Net by more than 10\%. Besides, when compared with another network, our method performs best, especially on correspondence classification task, by nearly 5 to 10 \% over the current best approach in terms of F1-measure. These experiments demonstrate
that the proposed network behaves favorably to the state-of-the-art approaches.
%It is worth noting that our network works better when using RANSAC as a post-processing method. 

\subsection{Ablation studies}

\textbf{HA Block.} To demonstrate the performance
of HA block, we replace the CN Block in the baseline CN-Net \cite{moo2018learning} with the HA block. Both the Bayesian attentive context normalization (BACN) and channel-wise attention (CA) are tested specifically as Tab. \ref{tab:Ablation}. As a comparison, we also replace CN Block with ACN Block \cite{sun2020acne} to train the network. Both of the BACN and CA achieve a better result than ACN, and HA block (BACN + CA) achieves an improvement of about 2\% over ACN on both correspondence classification  and relative pose estimation results.

\noindent\textbf{Guided Loss.} 
We then replace the original loss of CN-Net with our Guided Loss. As shown in Tab. \ref{tab:Ablation}, the proposed Guided Loss (CN-Net + BACN + CA + G-Loss) achieves a better performance over the original loss of CN-Net (CN-Net + BACN + CA) on the terms of F1-measure. It shows that the proposed Guided Loss can significantly improve the performance of the classification task. Meanwhile, there is a nearly 2\% improvement in the performance of the relative pose task by simply replacing the classification loss without modifying the rest of the network. This is because under the supervision of the  proposed Guided Loss, the precision and recall of the classification results are more balanced,  which is more conducive to the regression of the $E$ matrix.

\noindent\textbf{Cascade \textit{vs.} No Cascade.} 
In order to show the performance of the proposed cascade architecture, we first deepen the layers of CN-Net from 12 to 18 and test the result as comparison.
Meanwhile, we also train the proposed cascade network, which is also a 18-layer network.  As shown in Tab. \ref{tab:Ablation}, only increasing the number of network layers, the performance of the network is not significantly improved. The performance of cascade network with the same number of layers is significantly better than non-cascaded networks. It implies that using the Guided Loss in a coarse-to-fine cascade manner can significantly improve network performance.

\begin{table}%%If you do not follow these requirements, your paper will be subject to expensive reformatting and special handling fees that can easily exceed the extra page fee.
	\centering%
	\caption{\textbf{P}recision (\%), \textbf{R}ecall (\%), \textbf{F1}-measure (\%) and mAP (\%) $5^{\circ}$, $10^{\circ}$ (using \textbf{weighted eight-point/RANSAC} as post-processing) on the YFCC100M$\&$SUN3D dataset of different classification loss functions. F-Loss is using Fn-measure as objective function, while G-Loss is the proposed Guided Loss.} \label{tab:with_other_loss}%
	\resizebox{0.95\columnwidth}{!}
	{
		\begin{tabular}{cccccc}
			\hline
			& P & R & F1 & mAP $5^{\circ} $ & mAP $10^{\circ}$ \\
			%\midrule
			\hline
			CE-Loss & \textbf{63.21} & 46.33 & 48.67 & 10.12/26.31 & 24.85/39.92 \\
			IBCE-Loss & 37.23 & \textbf{73.21} & 47.08 & 15.12/33.11 & 31.87/43.47 \\ 
			Focal Loss & 70.67 & 41.44 & 49.67 & 11.32/27.65 & 28.94/41.43 \\
			F-Loss & 44.72 & 67.43 & 51.11 & 10.82/28.90 & 26.16/39.27 \\
			G-Loss & 50.62 & 66.20 & \textbf{56.44} & \textbf{18.52/34.83} & \textbf{33.41/45.64}
			\\ 
			\hline
		\end{tabular}
	}	
	
\end{table}%\usepackage{times}

%\begin{table}
%	\caption{Performance comparison on the YFCC100M$\&$SUN3D dataset of different classification loss functions. Fn-Loss ($n$ = 1, 2) is using Fn-measure as objective function, while Guided Fn ($n$ = 1, 2) is the proposed Guided Loss, which uses Fn-measure as a guidance.}
%	\label{tab:with_other_loss}
%	\footnotesize
%	\centering
%	\begin{tabular}{llll}
%		\toprule
%		& mAP $5^{\circ} $ & mAP $10^{\circ} $ & mAP $20^{\circ} $ \\
%		\midrule
%		CE-Loss & 10.12/26.31 & 17.85/35.92 & 32.14/51.92 \\
%		IB-CE-Loss \cite{deng2018pixellink} & 15.12/33.12 & 22.65/42.97 & 34.32/54.63 \\ 
%		Focal Loss \cite{lin2017focal} & 11.32/27.65 & 18.94/37.43 & 32.04/52.10 \\
%		F1-Loss \cite{zhao2019optimizing} & 9.82/26.90 & 16.16/38.27 & 26.83/53.32 \\ 
%		%\checkmark & \checkmark & \checkmark & & & 28.2/38.5 \\ \hline
%		F2-Loss \cite{zhao2019optimizing} & 8.64/28.72 & 14.07/39.28 & 23.84/51.25 \\ 
%		Guided F1 & 15.90/33.42 & 23.98/43.57 & 35.69/55.32
%		\\
%		
%		Guided F2 & \textbf{18.52/34.83} & \textbf{25.41/44.64} & \textbf{36.98/56.12}
%		\\ 
%		\bottomrule
%	\end{tabular}
%\end{table}

\subsection{Guided Loss \textit{vs.} another loss.}
In order to further verify the performance of the proposed Guided Loss, we record the training curves of the weight, precision and recall in Fig. \ref{training_curve}. As shown in Fig. \ref{training_curve} (a),  $\lambda$ in the Guided Loss is dynamically changed, while $\lambda$ in IB-CE-Loss is set to fixed value 0.5. As a result, the Guided Loss can achieve a balance between precision and recall, as shown in Fig. \ref{training_curve} (b). Meanwhile, when using F1-measure, which considers precision and recall equally, as the guidance, the gap between precision and recall is always small. And when using F2-measure, which is more bias towards recall, the recall is always higher than precision. It shows that the result of Guided Loss always accords with the guided Fn-measure, which verifies the effect of the guidance.

Meanwhile, we train the CN-Net \cite{moo2018learning} with the different loss functions \cite{deng2018pixellink,zhao2019optimizing,lin2017focal} and precision, recall, F1-measure and mAP under $5^{\circ}$, $10^{\circ}$ are reported in Tab. \ref{tab:with_other_loss}. As discussed in Introduction, when using Fn-measure as objective function, some relaxation has to be made and not all of the samples are utilized for back propagation. Therefore, Fn-Loss does not even perform as well as IB-CE-Loss. For the proposed Guided Loss, the network can achieve a better result than the other loss functions.
This is because the Guided Loss can maintain the advantages of IB-CE-Loss while achieving a balance between precision and recall.

\section{Conclusion}
%We present an end-to-end network for removing the wrong matches from the putative match set. In this network, we propose a novel Guided loss to establish the direct connection between loss and measurement, and an IA Block as feature extraction backbone to eliminate the impact of outlier on global context. We conduct extensive experiments to analyze the performance of each module and the overall network framework in detail. These experiments demonstrate that GLA-Net behaves favorably to the state-of-the-art approaches.

In this paper, we present a Guided Loss, which shows a new idea of loss designing. In the proposed Guided Loss, the network is expected to optimize the Fn-measure. Instead of directly using Fn-measure as objective function, we propose to use Fn-measure as the guidance and still adopt the form of cross entropy. Thus, we can maintain
the advantage of cross entropy loss while optimizing the Fn-measure. In other tasks,
the loss function and evaluation criteria may be different from ours, but the idea of using evaluation criteria to adjust objective function can be used to design more loss functions.
Besides, a hybrid attention (HA) block, including a Bayesian attentive context normalization and a channel-wise attention, is proposed for better extracting global context. The Guided Loss and HA Block are combined in a cascade network for two-view geometry tasks. Through extensive experiments, we demonstrate that our network can achieve the state-of-the-art performance on benchmark dataset.

\section{Acknowledgements}
This work was supported by the National Natural Science
Foundation of China under Grant 61772213, Grant
61991412 and Grant 91748204.

\section{Appendix}

\subsection{Proof of Algorithm 1}
\textbf{Theorem.} 
Given two equations as follows:
\begin{equation}\label{condition}
	\partial F_{X}/{\partial F_{Y}} = \partial l_{X}/{\partial l_{Y}},
\end{equation}
\begin{equation}\label{conclusion}
	dloss \cdot dFn \leq  0,
\end{equation}
where $\partial l_{X}$ and $\partial l_{Y}$ are the partial derivatives of loss ($loss$) with respect to $X$ and $Y$, and $\partial F_{X}$ and $\partial F_{Y}$ are the partial derivatives of Fn-measure ($Fn$) with respect to $X$ and $Y$. $dloss$ and $dFn$ are the derivatives of $loss$ and $Fn$.

Then, Eq. \ref{condition} is a sufficient condition of Eq. \ref{conclusion} .

\textbf{Proof.} As introduced in the paper, Fn-measure ($Fn$) and  loss ($loss$) are both the dependent variables of X and Y. Specifically, the form of loss is as follows: 
\begin{equation}\label{rewritten_loss}
	\begin{aligned}
		l=-(\lambda \frac{1}{N_{pos}} \sum_{i=1}^{N_{pos}}\log(y_{i}) &+ \mu  \frac{1}{N_{neg}} \sum_{j=1}^{N_{neg}}\log(1-y_{j})), \\
		s.t. \quad \ \lambda + \mu = 1, N_{pos} &+ N_{neg} = N 
	\end{aligned}
\end{equation}
where $N_{pos}$ and $N_{neg}$ are the number of positive and negative samples. We compute the average loss terms of $TP$, $TN$, $FP$ and $FN$ samples respectively, denoted as $l_{TP}, l_{TN}, l_{FP}, l_{FN}$. Then the $loss$ can be transformed as follows:
\begin{equation}\label{simplify_loss}
	\begin{aligned}
		l = & \lambda/N_{pos} \cdot \{X \cdot l_{FN} +
		(N_{pos} - X) \cdot l_{TP}\}  \\
		+ & \mu/N_{neg} \cdot \{Y \cdot l_{FP} + (N_{neg} - Y) \cdot l_{TN}\}.
	\end{aligned}  
\end{equation}
Since both the $TP$ and $FN$ samples belong to positive class (ground truth is positive), the loss term of each sample are computed by $-log(y_{i})$ in Eq. \ref{rewritten_loss} , where $y_{i}$ ($0 \leq y_i \leq 1$)  is the logit value. In fact, if the logit value of a positive sample (ground truth is positive) is greater than 0.5, then it is a $TP$ sample. And if the logit value is smaller than 0.5, it is a $FN$ sample. Obviously, since $-log(y_{i})$ is a monotone decreasing function, then the loss term of each $FN$ sample is greater than $TP$ sample. Thus, the average loss of $FN$ samples is greater than that of $TP$ samples, i.e.,  
\begin{equation}\label{constraint_loss_tmp1}
	l_{FN} > l_{TP}.
\end{equation}
Similarly, the the average loss of $FP$ samples is greater than that of $TN$ samples, i. e., 
\begin{equation}\label{constraint_loss_tmp2}
	l_{FP} > l_{TN}.
\end{equation}
Then, we can compute the partial derivatives of loss with respect to $X$ and $Y$ from Eq. \ref{simplify_loss}, as follows:
\begin{equation}\label{p_l_x}
	\begin{aligned}
		\partial l_{X} &= \lambda/N_{pos} \cdot (l_{FN} - l_{TP}), \\
		\partial l_{Y} &= \mu/N_{neg}\cdot (l_{FP} - l_{TN})
	\end{aligned}
\end{equation}
According to the constraints of Eq. \ref{constraint_loss_tmp1} and \ref{constraint_loss_tmp2}, we can obtain the following constraints:
\begin{equation}\label{constraint_loss}
	\partial l_{X} > 0 ,\partial l_{Y} > 0.
\end{equation}
We then perform the same operations on $Fn$ to obtain the constraints of $Fn$. Specifically, $Fn$ is related with precision ($P$) and recall ($R$), and $P$, $R$
are both related with $X$, $Y$, as follows:
\begin{equation}\label{Fn_PR}
	\begin{aligned}
		Fn = (1 + n^{2}) \cdot P & \cdot R / (n^{2} \cdot P + R),
	\end{aligned}  
\end{equation}
\begin{equation}\label{PR_XY}
	\begin{aligned}
		P &= (N_{pos} -X)/(N_{pos} - X  + Y), \\
		R &= (N_{pos}-X)/N_{pos}.
	\end{aligned}  
\end{equation}

According to the compound derivation formula, we can compute the $\partial F_{X}$ and $\partial F_{Y}$ as follows:
\begin{equation}\label{f_x}
	\begin{aligned}
		\partial F_{X} &= \partial F_{P} \cdot \partial P_{X} + \partial F_{R} \cdot \partial R_{X}, \\
		\partial F_{Y} &= \partial F_{P} \cdot \partial P_{Y} + \partial F_{R} \cdot \partial R_{Y}.
	\end{aligned}
\end{equation}
The $\partial F_{P}$, $\partial F_{R}$, $\partial P_{X}$, $\partial P_{Y}$ all can be computed from Eq. \ref{Fn_PR} and \ref{PR_XY}, as follows:
\begin{equation}\label{f_p_x}
	\begin{aligned}
		\partial F_{P} &= (1 + n^{2}) \cdot R^{2} / {(n^{2}P + R)^{2}} \geq 0, \\
		\partial F_{R} &= n^{2}(1 + n^{2}) \cdot P^{2} / {(n^{2}P + R)^{2}} \geq 0, \\
		\partial P_{X} &= -Y / {(N_{pos}-X+Y)^{2}} \leq 0, \\
		\partial P_{Y} &= -(N_{pos}-X) / {(N_{pos}-X+Y)^{2}} \leq 0, \\
		\partial R_{X} &= -1 / {N_{pos}} \leq 0, \\
		\partial R_{Y} &= 0,
	\end{aligned}
\end{equation}
We can easily obtain the following constraints of $\partial F_{X}$ and $\partial F_{Y}$ from Eq. \ref{f_x}, \ref{f_p_x}:
\begin{equation}\label{constraint_fn}
	\begin{aligned}
		\partial F_{X}\leq 0 , \\
		\partial F_{Y} \leq 0.
	\end{aligned}
\end{equation}

\begin{figure*}
	\centering
	\includegraphics[width=2\columnwidth]{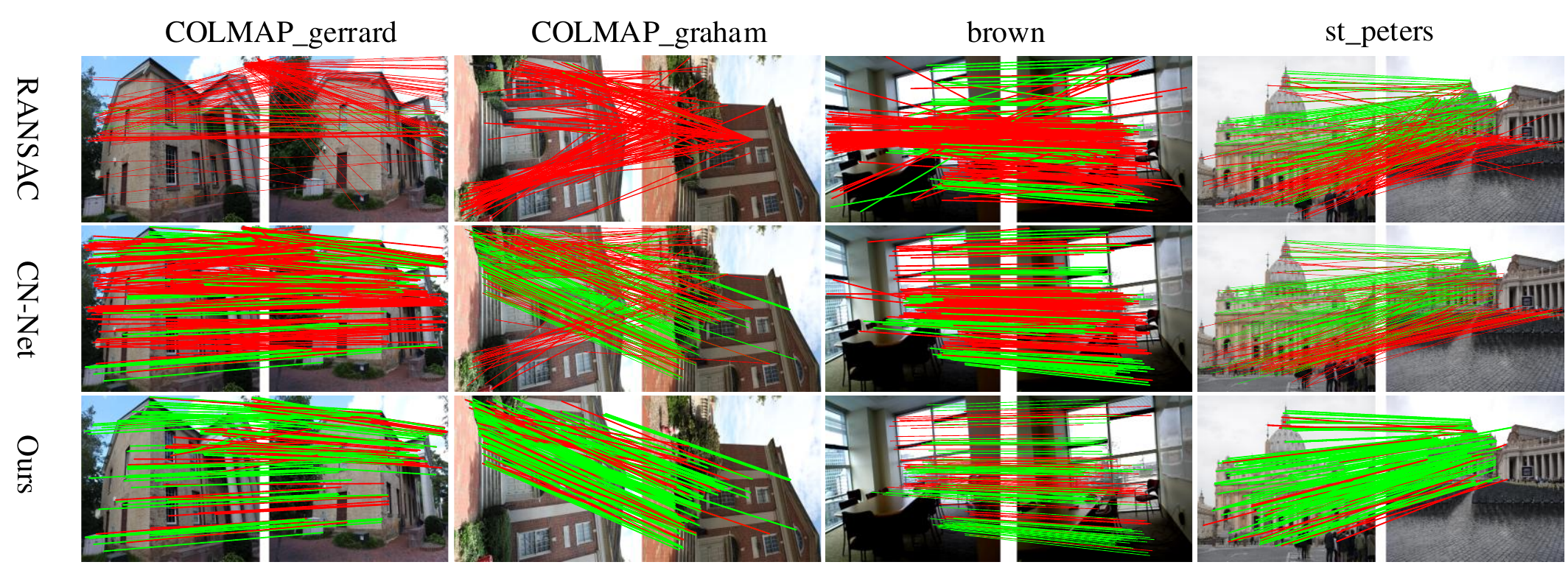} % Reduce the figure size so that it is slightly narrower than the column. Don't use precise values for figure width.This setup will avoid overfull boxes. 
	\caption{Visual comparison of matching results using RANSAC, CN-Net and our method. Images are taken from YFCC100M$\&$SUN3D and COLMAP datasets. Correspondences are in green if they are inliers, and in red otherwise. \textbf{Best viewed in color.}} 
	\label{visualize}
\end{figure*}

Meanwhile, since Fn-measure ($Fn$) and  loss ($loss$) are both the dependent variables of X and Y, the differential form of $loss$ and $Fn$ in Eq. \ref{conclusion} can be expressed as follows:
\begin{equation}\label{detail}
	\begin{aligned}
		dloss &= \partial l_{X} dX + \partial l_{Y} dY, \\
		dFn &= \partial F_{X} dX + \partial F_{Y} dY,
	\end{aligned}
\end{equation}
Thus, 
\begin{equation}\label{dldf_expand}
	\begin{aligned}
		&dloss \cdot dFn \\
		=&  (\partial l_{X} dX + \partial l_{Y} dY) \cdot (\partial F_{X} dX + \partial F_{Y} dY) \\
		=& \partial l_{X} \cdot \partial F_{X} \cdot (dX)^{2} + \partial l_{Y} \cdot \partial F_{Y} \cdot (dY)^{2} \\ 
		&+ (\partial l_{X} \cdot \partial F_{Y} + \partial l_{Y} \cdot \partial F_{X}) \cdot dX \cdot dY
	\end{aligned}
\end{equation}
According to the constraints of Eq. \ref{constraint_loss} and \ref{constraint_fn}, we can further expand Eq. \ref{dldf_expand} as follows:
\begin{equation}\label{dldf_expand2}
	\begin{aligned}
		&dloss \cdot dFn \\
		=& \partial l_{X} \cdot \partial F_{X} \cdot (dX)^{2} + \partial l_{Y} \cdot \partial F_{Y} \cdot (dY)^{2} \\
		&+ (\partial l_{X} \cdot \partial F_{Y} + \partial l_{Y} \cdot \partial F_{X}) \cdot dX \cdot dY \\ 
		= & \{\partial l_{X} \cdot \partial F_{X} \cdot (dX)^{2} + \partial l_{Y} \cdot \partial F_{Y} \cdot (dY)^{2} \\
		& - 2 \sqrt{\partial l_{X} \cdot \partial F_{X} \cdot \partial l_{Y} \cdot \partial F_{Y}} \cdot dX \cdot dY\} \\
		& + \{(\partial l_{X} \cdot \partial F_{Y} + \partial l_{Y} \cdot \partial F_{X}) \cdot dX \cdot dY \\
		& + 2 \sqrt{\partial l_{X} \cdot \partial F_{X} \cdot \partial l_{Y} \cdot \partial F_{Y}} \cdot dX \cdot dY\
		\} \\ 
		= & - (\sqrt{-\partial l_{X} \cdot \partial F_{X}} \cdot dX + \sqrt{-\partial l_{Y} \cdot \partial F_{Y}}\cdot dY)^{2} \\
		& - (\sqrt{-\partial l_{X} \cdot \partial F_{Y}} - \sqrt{-\partial l_{Y} \cdot \partial F_{X}})^{2} \cdot dX \cdot dY
	\end{aligned}
\end{equation}
If Eq. \ref{condition} holds, then, 
\begin{equation}\label{constraint4}
	\begin{aligned}
		\sqrt{-\partial l_{X} \cdot \partial F_{Y}} - \sqrt{-\partial l_{Y} \cdot \partial F_{X}} = 0,
	\end{aligned}
\end{equation}
then, 
\begin{equation}\label{constraint5}
	\begin{aligned}
		&dloss \cdot dFn \\ 
		=& - (\sqrt{-\partial l_{X} \cdot \partial F_{X}} \cdot dX + \sqrt{-\partial l_{Y} \cdot \partial F_{Y}}\cdot dY)^{2} \\ 
		\leq& 0
	\end{aligned}
\end{equation}

Thus, we can proof that Eq. \ref{condition} is a sufficient condition of Eq. \ref{conclusion}.

\subsection{Additional Experiments}

\textbf{Guided Loss with other baseline networks.}  We further analyze our Guided Loss by replacing the classification loss functions of other models with Guided Loss. We first train three recent networks, including CN-Net \cite{moo2018learning}, ACN-Net \cite{sun2019attentive} and NM-Net \cite{zhao2019nm}, with their original classification loss. Then we replace their classification loss with our Guided Loss. The results are reported in Tab. \ref{tab:overall_performance}. Each network with the supervision of our loss can lead to a result with more balanced precision and recall and a higher F1-measure. As a result, the each network increase the mAP by 1-3\% without modifying anything.

\begin{table}
	\caption{The results of three networks with both their original classification loss (without "+" in the table) and our F2-measure Guided Loss (with "+" in the table). \textbf{P}recision (\%), \textbf{R}ecall (\%), \textbf{F1}-measure (\%) and mAP (\%) $5^{\circ}$, $10^{\circ}$ (using \textbf{weighted eight-point/RANSAC} as post-processing) on the YFCC100M$\&$SUN3D dataset are reported.}
	\label{tab:overall_performance}
	\footnotesize
	\centering

	\resizebox{\columnwidth}{!}
	{
		\begin{tabular}{cccccc}
			\hline
			& P & R & F1 & mAP $5^{\circ} $ & mAP $10^{\circ}$ \\
			%\midrule
			\hline
			CN-Net & 37.23 & 73.21 & 47.08 & 15.12/33.11 & 31.87/43.47 \\
			CN-Net +  & 50.62 & 66.20 & 56.44 & 18.52/34.83 & 33.41/45.64\\ 
			ACN-Net & 40.20 & 80.01 & 50.48 & 25.87/35.68 & 35.66/46.04 \\
			ACN-Net + & 49.88 & 66.57 & 57.55 & 27.32/37.13 & 36.98/47.65 \\
			NM-Net & 40.66 & 71.66 & 50.74 & 17.70/34.09 & 32.80/42.92 \\
			NM-Net + & 46.72 & 65.44 & 55.68 & 19.89/36.79 & 34.32/44.02
			\\ 
			\hline
		\end{tabular}
	}
	%	\begin{tabular}{lllllll}
	%		\toprule
	%		& \multicolumn{3}{c}{$St\&Brown$} & \multicolumn{3}{c}{$Colmap$}  \\                 
	%		\cmidrule(r){2-4} \cmidrule(r){5-7}
	%		& mAP $5^{\circ} $ & mAP $10^{\circ} $ & mAP $20^{\circ} $  & mAP $5^{\circ} $ & mAP $10^{\circ} $ & mAP $20^{\circ} $ \\
	%		\midrule
	%		CN-Net \cite{moo2018learning} &  15.12/33.11
	%		& 31.87/43.47 & 43.62/54.81 & 11.82/26.89 & 18.44/30.82 & 23.89/34.52  \\
	%		CN-Net + & 18.53/34.80
	%		& 33.41/45.19 & 45.11/56.10 & 12.54/27.74 & 19.62/31.91 & 25.11/36.01  \\ 
	%		ACN-Net \cite{sun2019attentive}   & 25.87/35.68 & 35.66/46.04 & 47.69/58.25 & 21.65/30.40 & 25.72/34.89 & 30.02/41.58 \\
	%		ACN-Net + & 27.32/37.13 & 36.98/47.65 & 49.09/60.18 & 22.78/31.76 & 26.85/36.21 & 31.56/42.09 \\ 
	%		%\checkmark & \checkmark & \checkmark & & & 28.2/38.5 \\ \hline
	%		NM-Net \cite{zhao2019nm} & 17.70/34.09 & 32.80/42.92 & 43.74/54.62 & 20.96/31.72 & 23.08/33.42 & 29.18/39.62 \\ 
	%		
	%		NM-Net + & 19.89/36.79 & 34.32/44.02 & 44.85/56.17 & 21.78/32.43 & 24.05/34.87 & 30.22/40.74 
	%		\\ 
	%		
	%		\bottomrule
	%	\end{tabular}

\end{table}

\begin{table}
	\caption{The classification results of CN-Net \cite{moo2018learning} with different loss functions on YFCC100M$\&$SUN3D dataset. \textbf{G-Fn:} (n = 0.5, ..., 2.5) is the Guided Loss with the guidance of different Fn-measure. \textbf{IB-CE:} instance balance cross entropy loss. \textbf{CE:} cross entropy loss. The precision (P), recall (R) and 5 Fn-measure (Fn) are reported.} \label{Fn-score_guide}%
	\footnotesize
	\centering

	\resizebox{\columnwidth}{!}
	{
		\begin{tabular}{ccccccccc}
			\hline
			& P & R & F0.5 & F1 & F1.5 & F2 & F2.5 \\
			\hline
			G-F0.5 & \textbf{64.5} & 51.1 & \textbf{60.5} & 56.0 & 53.8 & 52.7 & 52.2 \\ 
			G-F1 & 56.8 & 57.8 & 57.2 & \textbf{57.3} & 55.9 & 56.7 & 56.6\\ 
			G-F1.5 & 48.6 & 64.6 & 50.6 & 54.3 & \textbf{58.2} & 59.6 & 60.9  \\ 
			G-F2 & 47.4 & 67.3 & 49.8 & 54.3 & 57.2 & \textbf{60.9} & 62.6 \\ 
			G-F2.5 & 43.6 & 69.2 & 46.6 & 52.2 & 57.1 & 60.5 & \textbf{63.8}  \\ 
			IB-CE & 37.2 & \textbf{77.1} & 40.0 & 47.1 & 53.8 & 59.4 & 62.9 \\ 
			CE & 63.2 & 48.3 & 56.7 & 52.1 & 51.1 & 50.7 & 50.5\\  
			\hline
		\end{tabular}
	}

\end{table}	

\textbf{Parameter $n$ of Guided Loss.} 
Our Guided Loss is designed to build a perfect negative correlation relationship between cross entropy loss and Fn-measure for better optimizing Fn-measure. To verify that our guidance actually works, we train the CN-Net \cite{moo2018learning} with different Fn-measure Guided Loss, respectively. Meanwhile, we also train the CN-Net with the cross entropy loss (CE-Loss) and Instance Balance cross entropy loss \cite{deng2018pixellink,moo2018learning} (IB-CE-Loss). The classification result are shown in Tab. \ref{Fn-score_guide}. We record the precision (P), recall (R), and five Fn-measures (n = 0.5, 1, 1.5, 2, 2.5 respectively) of the classification results on test dataset. As shown in Tab. \ref{Fn-score_guide}, with the guidance of a specific Fn-measure, our Guided Loss can achieve the best performance on this measurement. It shows the direct guidance of our loss. We can choose the corresponding Fn-measure to guide the loss according to the requirements of precision and recall of different tasks.

\textbf{Visual Result.} In order to show the classification result of our network, we visualize the matching results of RANSAC \cite{fischler1981random}, CN-Net \cite{moo2018learning} and our network in Fig. \ref{visualize}. 

\bibliography{egbib}
\bibliographystyle{aaai}

\end{document}